\documentclass{article}
\pdfoutput=1

\usepackage[preprint, nonatbib]{neurips_2022}

\usepackage[numbers]{natbib}
\usepackage{amssymb}
\usepackage[utf8]{inputenc} % allow utf-8 input
\usepackage[T1]{fontenc}    % use 8-bit T1 fonts
\usepackage{hyperref}       % hyperlinks
\usepackage{url}            % simple URL typesetting
\usepackage{booktabs}       % professional-quality tables
\usepackage{amsfonts}       % blackboard math symbols
\usepackage{nicefrac}       % compact symbols for 1/2, etc.
\usepackage{microtype}      % microtypography
\usepackage{xcolor}         % colors
\usepackage{tabu}         % colors
\usepackage{xspace,mfirstuc,tabulary}

\usepackage{amsmath}
\usepackage{multirow}
\usepackage{array, caption, floatrow, makecell, booktabs}
\usepackage{wrapfig}
\usepackage{floatrow}
\usepackage{comment}
\usepackage{footnote}
\usepackage{enumitem}
\usepackage{accents}

\usepackage{graphicx}
\usepackage{subfigure}
\usepackage{bbding}
\usepackage{ntheorem}
\usepackage{enumitem}
\usepackage{oplotsymbl}
\usepackage{multirow}
\usepackage{bbm}
\usepackage{adjustbox}

\usepackage{wrapfig}
\usepackage{comment}
\usepackage{enumitem}
\usepackage{accents}

\usepackage{tabu}

\usepackage{multirow,array}
\usepackage{color, colortbl}
\definecolor{Gray}{gray}{0.93}

\usepackage{booktabs}
\usepackage{pifont}
\newcommand{\cmark}{\ding{51}}%
\newcommand{\xmark}{\ding{55}}%
\newlength\savewidth\newcommand\shline{\noalign{\global\savewidth\arrayrulewidth
  \global\arrayrulewidth 1pt}\hline\noalign{\global\arrayrulewidth\savewidth}}
\newcommand\paperurl[1]{{\footnotesize{\color{blue}{\url{#1}}}}}

\title{A Strong and Reproducible Object Detector \\ with Only Public Datasets}

\author{%
  \textbf{Tianhe Ren}\textsuperscript{\rm 1 $\clubsuit$},\; \textbf{Jianwei Yang}\textsuperscript{\rm 2 $\clubsuit$},\; \textbf{Shilong Liu}\textsuperscript{\rm 1 $\clubsuit$},\; \textbf{Ailing Zeng}\textsuperscript{\rm 1 $\clubsuit$}, \\
  \textbf{Feng Li}\textsuperscript{\rm 1},\; \textbf{Hao Zhang}\textsuperscript{\rm 1},\; \textbf{Hongyang Li}\textsuperscript{\rm 1},\; \textbf{Zhaoyang Zeng}\textsuperscript{\rm 1},\; \textbf{Lei Zhang}\textsuperscript{\rm 1 $\spadesuit$}.
  \vspace{0.2cm} \\
    \textsuperscript{\rm 1} International Digital Economy Academy (IDEA)\\
    \textsuperscript{\rm 2} Microsoft Research, Redmond  \\
    $\clubsuit$ Equal Contribution. List in random. $\spadesuit$ Project Lead \\\\
    Code will be available at: \\
    \url{https://github.com/microsoft/FocalNet} \\
    \url{https://github.com/IDEA-Research/Stable-DINO}
}
\begin{document}

\maketitle

%%%%% Abstract
% \begin{abstract}
%   We investigated the limitation of the large object detection model on publicly available datasets, and present a robust detector named Focal-Stable-DINO. We built our model by combining a strong backbone \textit{FocalNet-Huge}~\cite{yang2022focal} and an efficient Stable-DINO~\cite{stable-dino} detector. We pretrained our model on Object365 and finetuned it on COCO dataset, both of which are public dataset. Our model achieved a remarkable \textbf{64.6} mAP on COCO minimal and \textbf{xxx} mAP on COCO test-dev, setting a new state-of-the-art performance among detectors trained exclusively on public datasets.
% \end{abstract}

% 所有的东西都是reachable的，告诉大家如何得到这个number
% simple, strong, 
% public data
% available
% sota

\begin{abstract}
  % With the recent developments in large-scale models, Object Detection has a surge of remarkable outcomes. Many works have attempted to achieve new state-of-the-art performance by utilizing: (1) backbones with massive parameters, (2) more training data including private datasets, (3) rich pre-training techniques such as masked image modeling (MIM) for pre-training. In this report, we investigate the limitation of large object detection models on publicly available datasets. We first present a strong detector, named \textbf{Focal-Stable-DINO}, by combining the powerful backbone, FocalNet-Huge~\cite{yang2022focal}, with the effective Stable-DINO detector~\cite{liu2023detection}. Our model was pre-trained on the Object365 dataset and fine-tuned on the COCO dataset, both of which are publicly available datasets. Notably, without Test Time Augmentation, our model achieved a remarkable \textbf{64.6AP} on COCO \texttt{minival} and \textbf{64.8AP} on COCO \texttt{test-dev} using only 700M parameters. Subsequently, we provide our insights and prospects regarding the development of object detection from our perspective.

% This work presents a strong and reproducible object detection model \textbf{Focal-Stable-DINO}, achieving an impressive performance \textbf{64.6 AP} on the COCO \texttt{minival} dataset and \textbf{64.8 AP} on COCO \texttt{test-dev} using only 700M parameters without any test time augmentation. 
This work presents \textbf{Focal-Stable-DINO}, a strong and reproducible object detection model which achieves \textbf{64.6 AP} on COCO \texttt{val2017} and \textbf{64.8 AP} on COCO \texttt{test-dev} using only 700M parameters \emph{without} any test time augmentation. It explores the combination of the powerful FocalNet-Huge backbone~\cite{yang2022focal} with the effective Stable-DINO detector~\cite{liu2023detection}. Different from existing SOTA models that utilize an extensive number of parameters and complex training techniques on large-scale private data or merged data, our model is exclusively trained on the publicly available dataset Objects365~\cite{shao2019objects365}, which ensures the reproducibility of our approach. 

  \begin{figure}[h]
    \centering
    % \vspace{-0.2cm}
    \includegraphics[width=1.0\linewidth, bb=0 0 800 400]{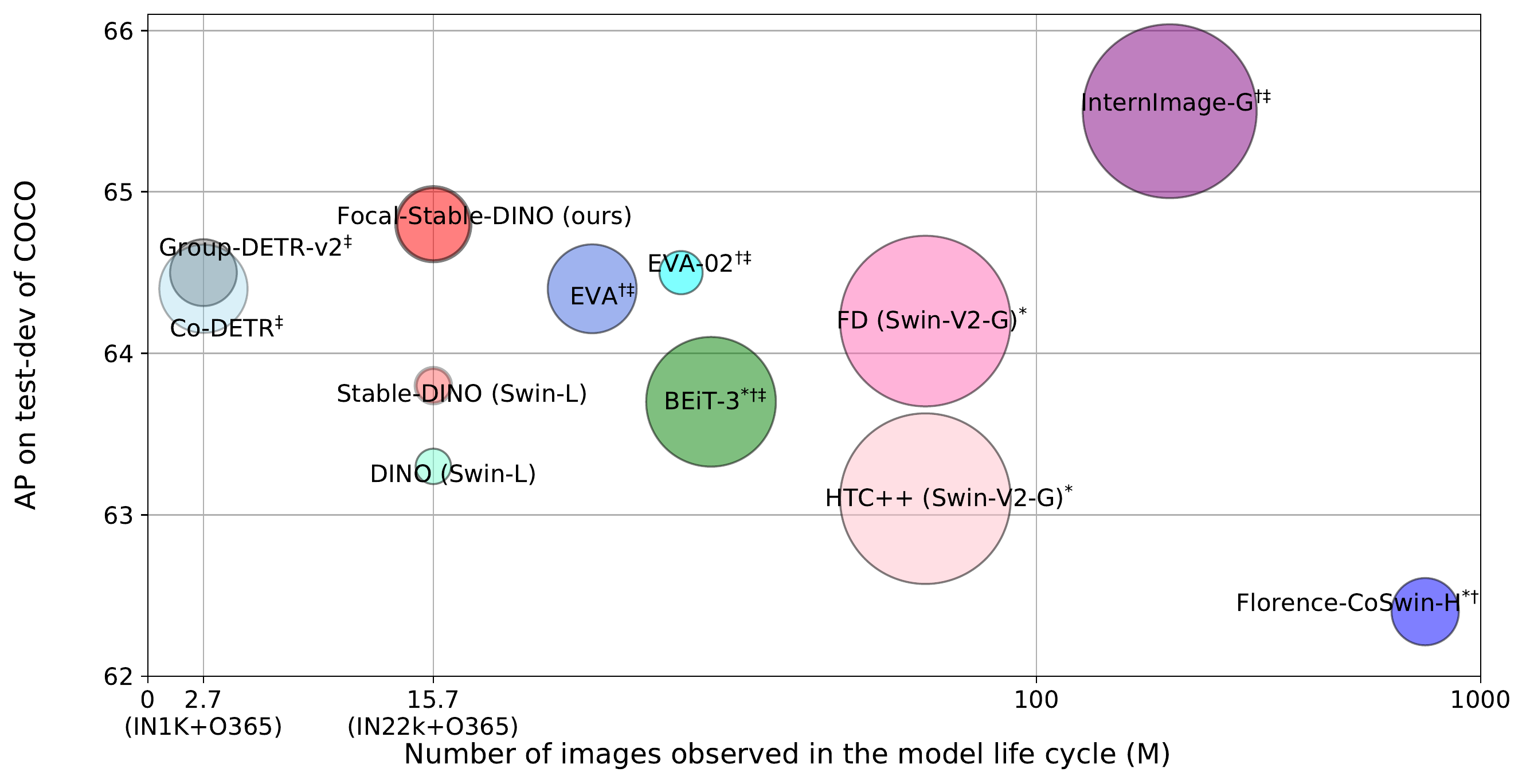}
    \vspace{-0.2cm}
    \caption{Comparison of the proposed \textbf{Focal-Stable-DINO} with existing SOTA models on COCO \texttt{test-dev}. Circle size indicates the model size. We use superscripts $^{*}$, $^{\dag}$, and $^{\ddag}$ to denote the models with private data, usage of large-scale image-text pairs, 
    and sophisticated training methods like masked image training, respectively.
    % and masked image training.
    % "\#" means using TTA (Test Time Augmentation)
    }
    \label{fig:coco_sota_testdev}
    % \vspace{-0.4cm}
\end{figure}

\end{abstract}

%%%%% Body Text
\section{Introduction}
Object detection has undergone significant development in recent years, primarily driven by improvements in model structure, data scale, and training strategies. At the model level, various studies, such as \cite{gao2021fast, ren2015faster, liu2022dabdetr, li2022dn, zhang2022dino, stabledino, hu2023yoso}, have focused on enhancing model designs for better results. Similarly, advanced training strategies, such as masked image modeling \cite{mae, chen2022group} and image-text contrastive learning~\cite{yuan2021florence, yang2022unified, fang2022eva}, have been adopted to improve the model representations.

After reaching a plateau in model design and training strategy, recent works have begun to explore using larger pre-trained models and more extensive data volumes, particularly private datasets, to continue enhancing the model's performance. 
For example, works such as EVA~\cite{fang2022eva, fang2023eva}, BEiT~\cite{wang2022image}, and InternImage-DINO~\cite{wang2022internimage} have utilized extraordinarily large backbones and ample training data to achieve new state-of-the-art results on COCO 2017 \texttt{val2017} and \texttt{test-dev}.
Unfortunately, the progress of these advancements has been hindered by the limited accessibility of the private training data, making replication of the reported strong results challenging. In addition, the complex data curation pipeline (\textit{e.g.}, merging multiple datasets) and sophisticatedly designed pre-training methods (\textit{e.g.}, masked image modeling and contrastive learning) have made these techniques hard to reproduce for researchers without access to private data and code.

In object detection, a detector typically comprises two primary components - a backbone and a detection head. Our strategy involves assembling publicly available resources in a straightforward pipeline. Recently, FocalNet-Huge~\cite{yang2022focal} with 689M parameters is a publicly available model that is pre-trained only on ImageNet-22K \cite{deng2009imagenet} and exhibits strong transferability to object detection. Stable-DINO~\cite{liu2023detection} proposes a position-supervised loss and a position-modulated cost to relieve the matching stability problem across different decoder layers in DINO~\cite{zhang2022dino}, making only one optimization path available and thus mitigating the multi-optimization paths issue.
Based on the above analyses, we develop a strong object detector, named \textbf{Focal-Stable-DINO}, without relying on complex training methods like masked image modeling or incorporation of private data or image-text pairs. 
Focal-Stable-DINO is not only strong but also reproducible, making it a compelling model for the research community to continually make improvements. This is due to its foundation on a powerful Focal-Huge backbone and the incorporation of a high performance Stable-DINO detector.
Stable-DINO with a Swin-L backbone achieves an impressive average precision (AP) of 63.8 on the COCO \texttt{test-dev} dataset.
By replacing the Swin-L backbone with a Focal-Huge backbone, Focal-Stable-DINO can be further improved and achieves \textbf{64.6 AP} on COCO \texttt{val2017} and \textbf{64.8 AP} on COCO \texttt{test-dev}, without any testing techniques such as test time augmentation.

\section{Method}

% \paragraph{Public Available Resources}
\paragraph{Architecture.} We adapt FocalNet-Huge~\cite{yang2022focal} as our backbone, which exhibits an exceptional capability among large models with model parameters of only 689M. We exploit Stable-DINO~\cite{liu2023detection} as our detector. Stable-DINO proposes a {position-supervised loss} which can significantly improve the training stability of DETR-variants and has also yielded remarkable results on large models.

All resources for our final results are publicly accessible, which are summarized in Table \ref{tab:pub_resource}.

\begin{table*}[ht]
\centering\setlength{\tabcolsep}{7pt}
\renewcommand{\arraystretch}{1.5}
%\footnotesize
\vspace{-0mm}
\resizebox{0.9\columnwidth}{!}{%
\begin{tabular}{c|c|l}
\shline
Resource & Usage For & Link  \\
\shline
FocalNet-Huge \cite{yang2022focal} & Backbone & \url{https://github.com/microsoft/FocalNet} \\
Stable-DINO \cite{liu2023detection} & Detection Head & \url{https://github.com/IDEA-Research/StableDINO} \\
\hline
Objects365 \cite{shao2019objects365} & Detection Pre-train & \url{https://www.objects365.org/} \\
COCO \cite{lin2014microsoft} & Detection Fine-tune & \url{https://cocodataset.org/} \\
\shline
\end{tabular}
\vspace{-0.2cm}
\caption{{Public reachable resources we used.}}
\label{tab:pub_resource}}
\end{table*}

\begin{table*}[ht]
\centering\setlength{\tabcolsep}{7pt}
\renewcommand{\arraystretch}{1.5}
%\footnotesize
%\vspace{-0mm}
\resizebox{1.0\columnwidth}{!}{%
\begin{tabular}{cccccccc}
\shline
\textbf{Method} & \textbf{Backbone} & \#\textbf{Param.} & \textbf{Pre-training Dataset} & \textbf{TTA} &  \textbf{w/ Mask} & \texttt{val} \textbf{AP} & \texttt{test} \textbf{AP} \\
\shline
\multirow{1}[1]{*}{\makecell{Soft-Teacher~\cite{xu2021end}}} & \multirow{1}[1]{*}{\makecell{Swin-L}}  & \multirow{1}[1]{*}{\makecell{284M}}  & \multirow{1}[1]{*}{\makecell{IN22k + O365 + COCO(unlabeled)}} & \multirow{1}[1]{*}{\makecell{\cmark}} & \multirow{1}[1]{*}{\makecell{\cmark}} & \multirow{1}[1]{*}{\makecell{60.7}} & \multirow{1}[1]{*}{\makecell{61.3}} \\
\multirow{1}[1]{*}{\makecell{Florence~\cite{yuan2021florence}}} & \multirow{1}[1]{*}{\makecell{CoSwin-H}}  & \multirow{1}[1]{*}{\makecell{637M}}  & \multirow{1}[1]{*}{\makecell{FLD900M + merged data$^a$}}  & \multirow{1}[1]{*}{\makecell{\cmark}} & \multirow{1}[1]{*}{\makecell{\xmark}} & \multirow{1}[1]{*}{\makecell{62.0}} & \multirow{1}[1]{*}{\makecell{62.4}} \\
HTC++~\cite{chen2019hybrid} & SwinV2-G~\cite{liu2022swin}  & 3.0B  & IN-22K-ext + O365 & \cmark & \cmark & 62.5 & 63.1 \\
DINO~\cite{zhang2022dino} & Swin-L~\cite{liu2021Swin}  & 218M  & IN-22K + O365 & \cmark & \xmark & 63.2 & 63.3 \\
\multirow{1}[1]{*}{\makecell{BEiT-3~\cite{wang2022image}}} & \multirow{1}[1]{*}{\makecell{ViT-g~\cite{zhai2022scaling}}}  & \multirow{1}[1]{*}{\makecell{1.9B}}  & \multirow{1}[1]{*}{\makecell{merged data$^b$ + O365}} & \multirow{1}[1]{*}{\makecell{\cmark}} & \multirow{1}[1]{*}{\makecell{\cmark}} & \multirow{1}[1]{*}{\makecell{-}}& \multirow{1}[1]{*}{\makecell{63.7}} \\
FD~\cite{wei2022contrastive} & SwinV2-G & 3.0B  & IN-22K-ext + O365 & \cmark & \cmark & - & 64.2 \\
Stable-DINO~\cite{liu2023detection} & Swin-L & 218M  & IN-22K + O365 & \xmark & \xmark & {63.7} & 63.8 \\
FocalNet-DINO~\cite{yang2022focal} & FocalNet-Huge~\cite{yang2022focal} & 689M  & IN-22K + O365 & \xmark & \xmark & 64.0 & - \\
FocalNet-DINO~\cite{yang2022focal} & FocalNet-Huge~\cite{yang2022focal} & 689M  & IN-22K + O365 & \cmark & \xmark & 64.2 & 64.4 \\
Group-DETR-v2~\cite{chen2022group} & ViT-Huge~\cite{dosovitskiy2020image} & 629M  & IN-1K + O365 & \cmark & \xmark & - & 64.5 \\
Co-Deformable-DETR~\cite{zong2022detrs} & MixMIM-g~\cite{liu2022mixmim} & 1.0B  & IN-1K + O365 & \cmark & \xmark & 64.4 & 64.5 \\
Internimage-DINO~\cite{wang2022internimage} & InternImage-XL & 602M  & IN-22K + O365 & \cmark & \xmark & 64.2 & 64.3 \\

\multirow{1}[1]{*}{\makecell{EVA-01}}~\cite{fang2022eva} & \multirow{1}[1]{*}{\makecell{EVA}} & \multirow{1}[1]{*}{\makecell{1.0B}}  & \multirow{1}[1]{*}{\makecell{merged-30M}} & \multirow{1}[1]{*}{\makecell{\xmark}} &\multirow{1}[1]{*}{\makecell{\cmark}} & \multirow{1}[1]{*}{\makecell{64.2}} & \multirow{1}[1]{*}{\makecell{64.4}}\\

\multirow{1}[1]{*}{\makecell{EVA-01~\cite{fang2022eva}}} & \multirow{1}[1]{*}{\makecell{EVA}} & \multirow{1}[1]{*}{\makecell{1.0B}}  & \multirow{1}[1]{*}{\makecell{merged-30M}} & \multirow{1}[1]{*}{\makecell{\cmark}} & \multirow{1}[1]{*}{\makecell{\cmark}} & \multirow{1}[1]{*}{\makecell{64.5}} & \multirow{1}[1]{*}{\makecell{64.7}}\\

\multirow{1}[1]{*}{\makecell{EVA-02}}~\cite{fang2023eva} & \multirow{1}[1]{*}{\makecell{EVA-02}} & \multirow{1}[1]{*}{\makecell{304M}}  & \multirow{1}[1]{*}{\makecell{merged-38M}} & \multirow{1}[1]{*}{\makecell{\xmark}} & \multirow{1}[1]{*}{\makecell{\cmark}} & \multirow{1}[1]{*}{\makecell{64.1}} & \multirow{1}[1]{*}{\makecell{64.5}}\\
\shline
\rowcolor{gray!15} Focal-Stable-DINO & FocalNet-Huge~\cite{yang2022focal} & 689M  & IN-22K + O365 & \xmark & \xmark & \textbf{64.6} & \textbf{64.8} \\
\shline
\end{tabular}
\vspace{-0.2cm}
\caption{{Comparison to the state-of-the-art models on COCO \texttt{val2017} (\texttt{val}) and \texttt{test-dev} (\texttt{test}) trained with only publicly available datasets. "TTA" means using test time augmentation in testing. "w/ Mask" means using mask annotations when finetuning the detectors. Details of the used dataset are described as follows:\\
\tiny{"IN-1K": ImageNet-1K, 1M image data in total\\
"IN-22K": ImageNet-22K, 14M image data in total\\
"IN-22K-ext": ImageNet-22K-ext, 70M private image data in total\\ 
"O365": Objects365, 1.7M image data in total\\ 
"merged data$^a$": FourODs + INBoxes+ GoldG + CC15M + SBU\\ 
"merged data$^b$ ": IN-22K (14M) + Image-Text (35M) + Text (160GB)\\ 
"merged-30M": IN-22K (14M) + CC12M + CC3M + COCO + ADE20K + Objects365\\
"merged-38M": IN-22K (14M) + CC12M + CC3M + COCO + ADE20K + Objects365 + OpenImage}
}}
\label{tab:benchmark_comparison}}
\end{table*}

\section{Experiments}

  \paragraph{Implementation details.} We pre-train our detector on Objects365~\cite{shao2019objects365} and fine-tune it on COCO~\cite{lin2015microsoft}. We adopt the same hyper-parameters as DINO \cite{zhang2022dino} and FocalNet \cite{yang2022focal}, including utilizing $1.5 \times$ resolution during fine-tuning, using 1000 De-noising queries, and a relatively small noise ratio~\cite{li2022dn}. We set a larger classification weight of $6.0$ during loss computation as Stable-DINO \cite{liu2023detection} for the position-supervised loss.

  \paragraph{Comparison with state-of-the-art models on COCO.} We compare our model with previous SOTA results in Table~\ref{tab:benchmark_comparison}. When trained exclusively on publicly available datasets and using only box annotations, with the Focal-Huge backbone, our model achieves \textbf{64.6 AP} on COCO \texttt{val2017} without utilizing any test time augmentation. Furthermore, on COCO \texttt{test-dev}, Focal-Stable-DINO achieves \textbf{64.8 AP} without employing any test time augmentation.

  \paragraph{Main results on COCO} We report our main results on COCO \textit{val2017} and \textit{test-dev} in Table~\ref{tab:sota_minival} and Table~\ref{tab:sota_test_dev}, respectively. Among models trained with only public datasets, Focal-Stable-DINO achieves the best result on COCO. 

  \begin{table*}[ht]
\centering\setlength{\tabcolsep}{7pt}
\renewcommand{\arraystretch}{1.5}
\footnotesize
\vspace{-0mm}
\resizebox{1.0\columnwidth}{!}{%
\begin{tabular}{cccc|cccccc}
\shline
Method & Backbone & \#Params. & TTA? & AP & AP$_{50}$ & AP$_{75}$ & AP$_{S}$ & AP$_{M}$ & AP$_{L}$ \\
\shline
% Swin-Stable-DINO & Swin-Large  & 218M & \xmark  & 63.7 & 80.7 & 70.3 & 49.7 & 67.6 & 78.1 \\
% Swin-Stable-DINO & Swin-Large  & 218M & \cmark  &  &  &  &  &  &  \\
Focal-Stable-DINO & Focal-Huge  & 689M & \xmark  & {64.6} & 81.5 & 71.4 & 50.4 & 68.5 & 78.5 \\
% Focal-Stable-DINO & Focal-Huge  & 689M & \cmark  &  &  &  & & & \\
\shline
\end{tabular}
\vspace{-0.2cm}
\caption{{Our Focal-Stable-DINO results on COCO \texttt{val2017} trained only \textit{on publicly available datasets} without utilizing any test time augmentation techniques.}}
\label{tab:sota_minival}}
\end{table*}

  \begin{table*}[ht]
\centering\setlength{\tabcolsep}{7pt}
\renewcommand{\arraystretch}{1.5}
\footnotesize
\vspace{-0mm}
\resizebox{1.0\columnwidth}{!}{%
\begin{tabular}{cccc|cccccc}
\shline
Method & Backbone & \#params & TTA? & AP & AP$_{50}$ & AP$_{75}$ & AP$_{S}$ & AP$_{M}$ & AP$_{L}$ \\
\shline
% Swin-Stable-DINO & Swin-Large  & 218M & \xmark  & 63.8 & 80.7 & 70.3 & 47.4 & 66.8 & 77.3 \\
% Swin-Stable-DINO & Swin-Large  & 218M & \cmark  &  &  &  &  &  &  \\
Focal-Stable-DINO & Focal-Huge  & 689M & \xmark  & 64.8 & 81.7 & 71.5 & 48.6 & 67.6 & 78.0 \\
% Focal-Stable-DINO & Focal-Huge  & 689M & \cmark  &  &  &  & & & \\
\shline
\end{tabular}
\vspace{-0.2cm}
\caption{{Our Focal-Stable-DINO results on COCO \texttt{test-dev} trained only \textit{on publicly available datasets} without utilizing any test time augmentation techniques.}}
\label{tab:sota_test_dev}}
\end{table*}

% \section{Analysis and Prospects}
\section{Analysis}
\subsection{Analysis of Model Prediction Quality}
\begin{figure}[h]
    \centering
    % \vspace{-0.2cm}
    \includegraphics[width=1.0\linewidth, bb=0 0 1000 350]{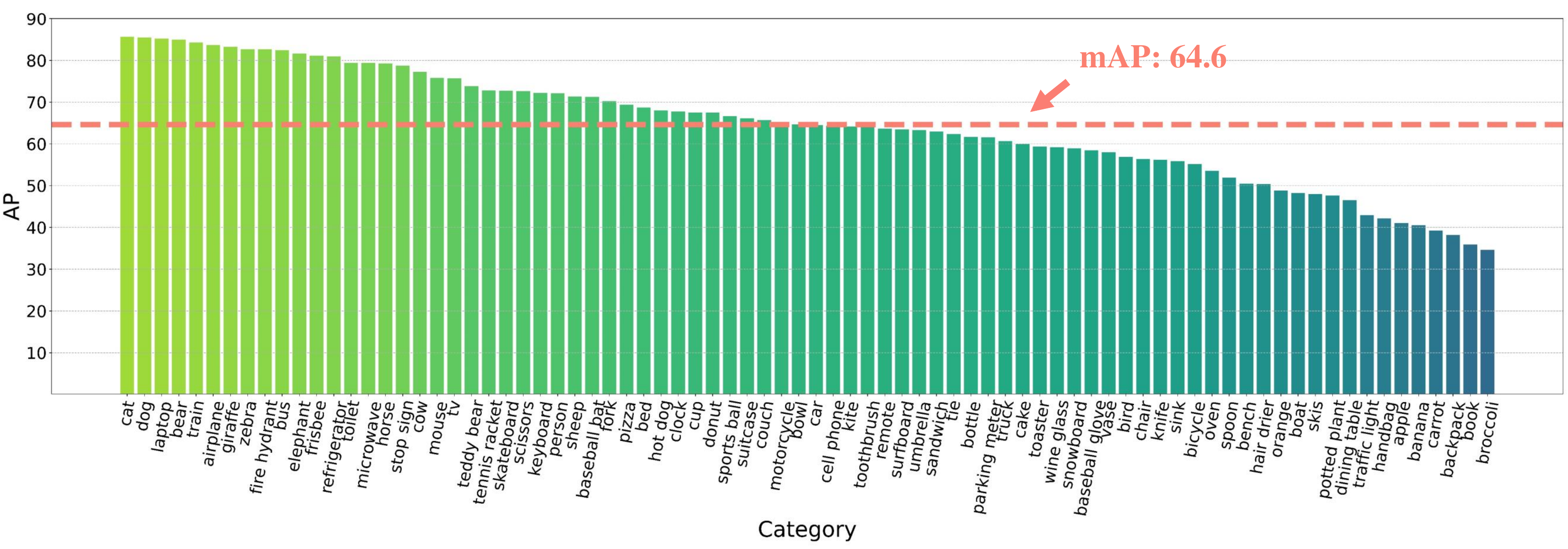}
    \vspace{-0.2cm}
    \caption{AP for each category of COCO \texttt{val2017}.
    }
    \label{fig:ap_per_category}
    % \vspace{-0.4cm}
\end{figure}

We visualize the average precision (AP) values for each category in our model's predictions. From Fig.~\ref{fig:ap_per_category}, we observe that there is still room for improvement in the performance of many categories below the line of 64.6 mAP.
% \vspace{-0.3cm}
% \vspace{-0.5cm}

\begin{figure}[h]
    \centering
    % \vspace{-0.2cm}
    \includegraphics[width=1.0\linewidth, bb=0 0 1200 500]{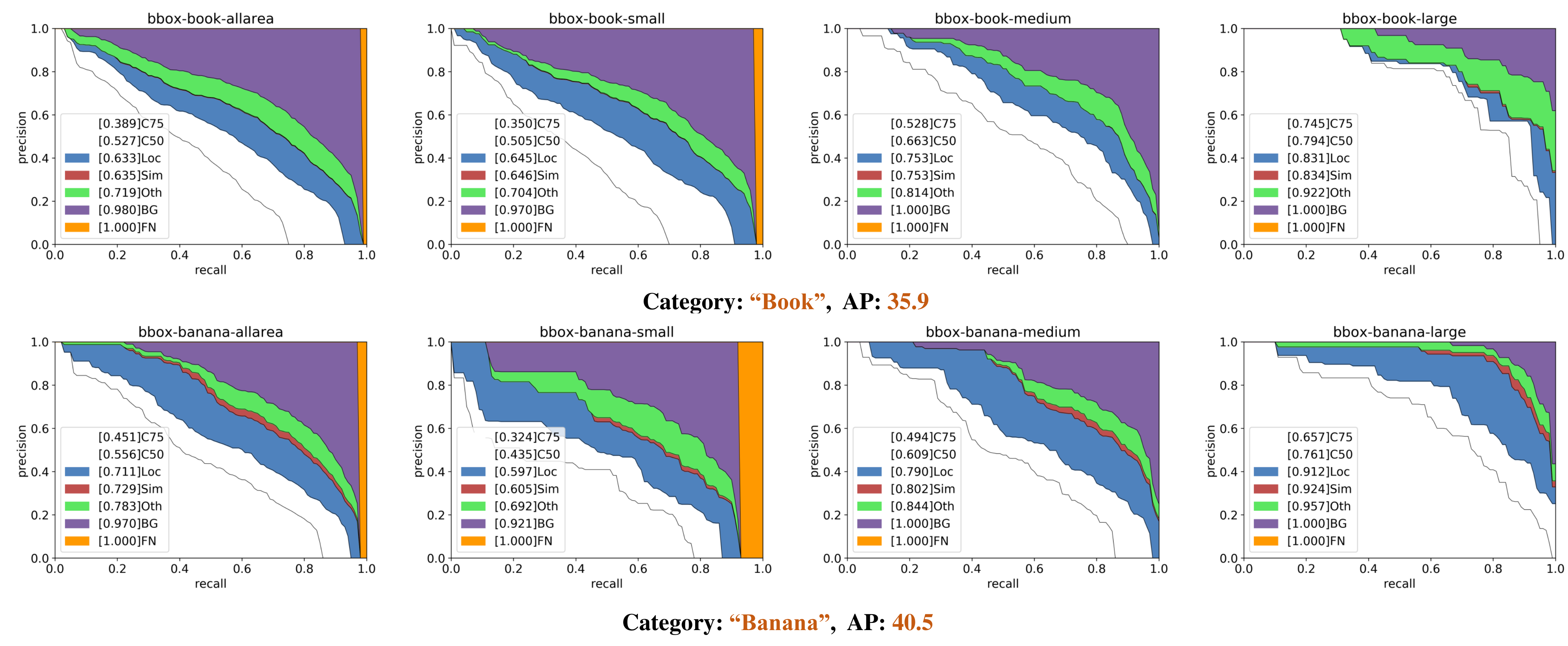}
    % \vspace{-0.1cm}
    \caption{The precision-recall curves for two bad cases.
    }
    \label{fig:lower_ap_case}
    % \vspace{-0.4cm}
\end{figure}

\begin{figure}[ht]
    \centering
    % \vspace{-0.2cm}
    \includegraphics[width=1.0\linewidth, bb=0 0 1200 500]{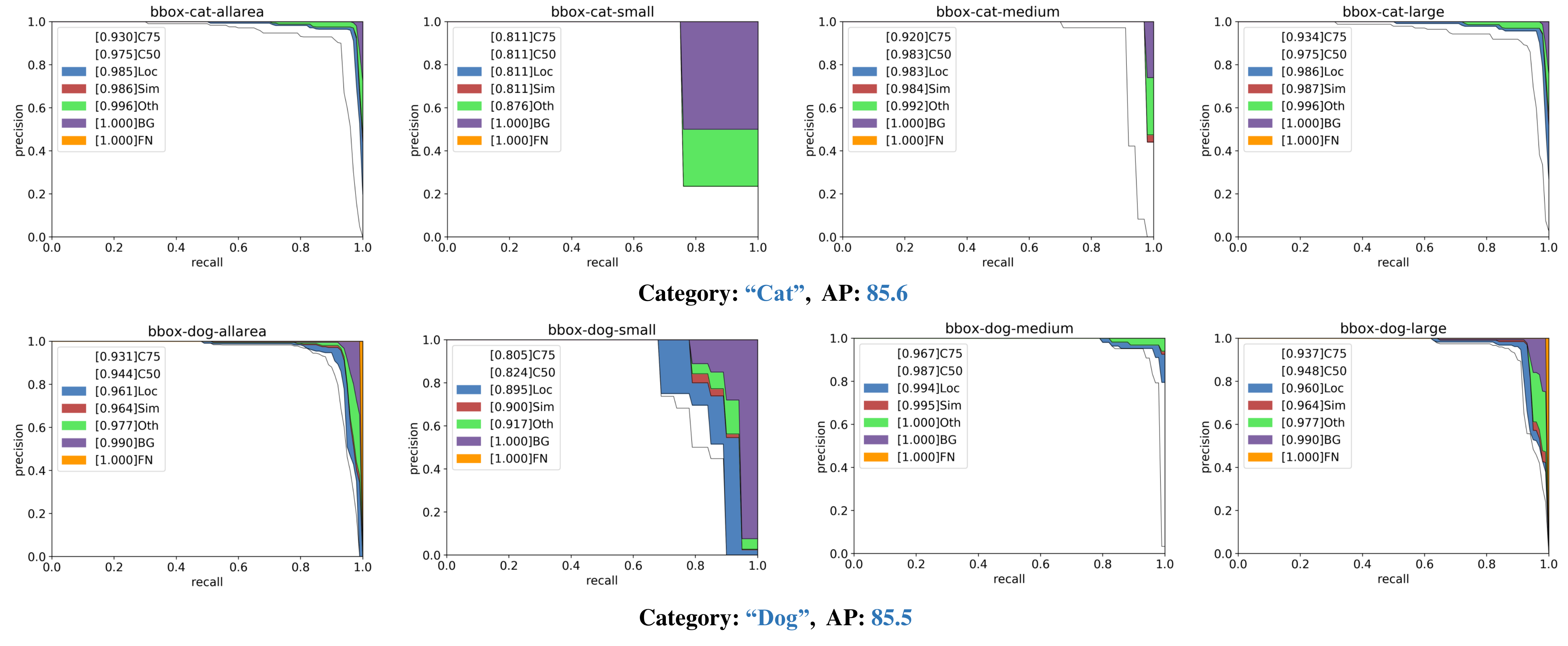}
    % \vspace{-0.1cm}
    \caption{The precision-recall curves for two good cases.
    }
    \label{fig:higher_ap_case}
    % \vspace{-0.4cm}
\end{figure}

Inspired by~\cite{hoiem2012diagnosing}, we further conduct a detailed analysis of the classes with low and high AP values and show their precision-recall curves in Fig.~\ref{fig:lower_ap_case} and Fig.~\ref{fig:higher_ap_case}. As shown in Fig.~\ref{fig:lower_ap_case}, correcting the localization errors in the detection results of the "Book" category increases the AP from 0.389 to 0.633. Furthermore, removing the background false positives further increases the AP to 0.980. Similar improvements can be observed in the "Banana" category, where fixing the background false positives leads to an AP of 0.970, which indicates that the prediction errors in these categories are mainly due to background confusion. As shown in Fig.~\ref{fig:higher_ap_case}, for categories with higher AP, correcting the localization errors results in a greater improvement than correcting the background confusion. Additionally, there is still ample room for improvement in the localization of small objects.


\begin{figure}[h]
    \centering
    % \vspace{-0.2cm}
    \includegraphics[width=0.8\linewidth, bb=0 0 600 400]{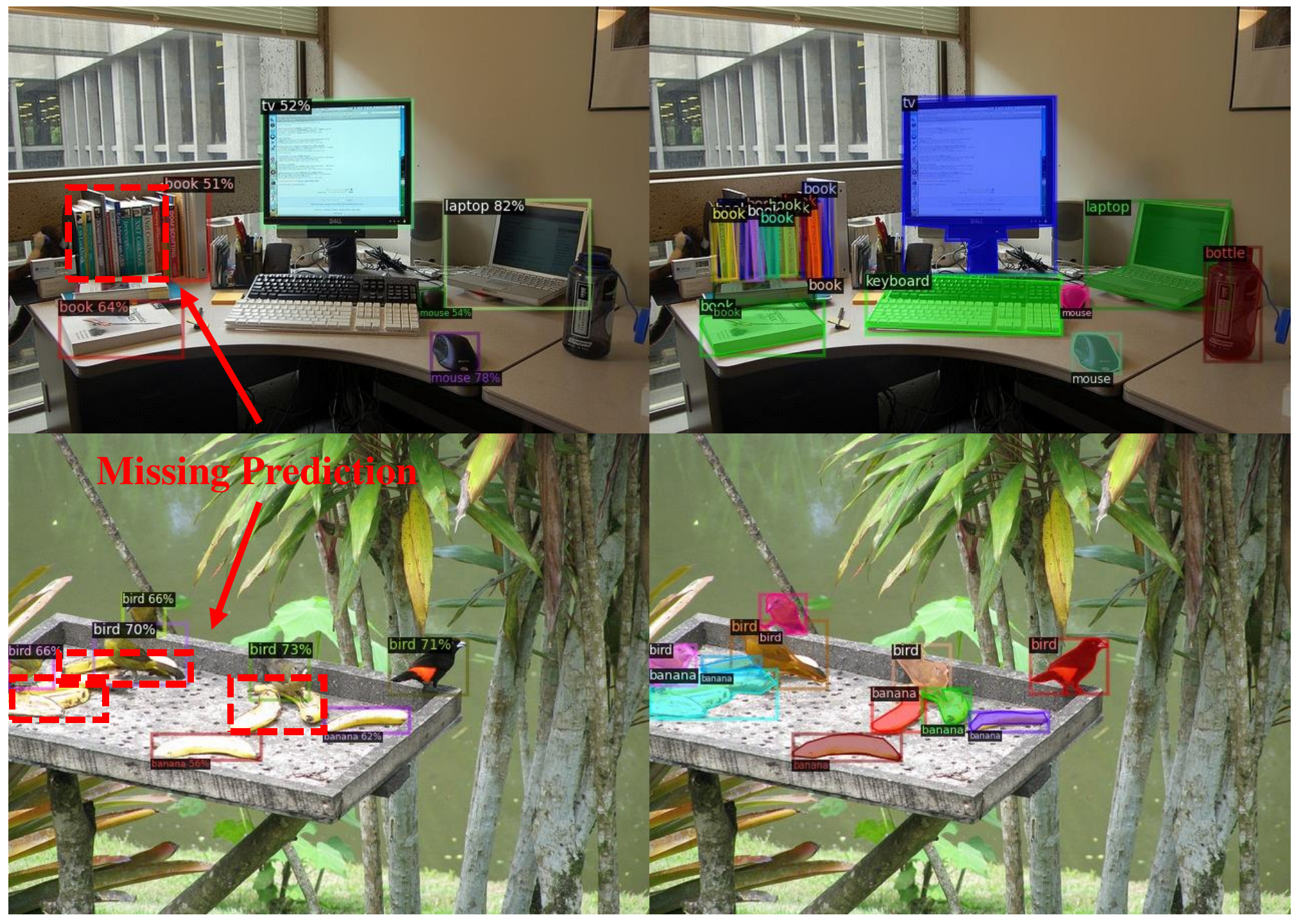}
    % \vspace{-0.1cm}
    \caption{Missing predictions for category "book" and "banana" (COCO Image ID: "000000026564" and "000000227187")
    }
    \label{fig:missing_prediction}
    % \vspace{-0.4cm}
\end{figure}

\begin{figure}[h]
    \centering
    % \vspace{-0.2cm}
    \includegraphics[width=0.8\linewidth, bb=0 0 650 400]{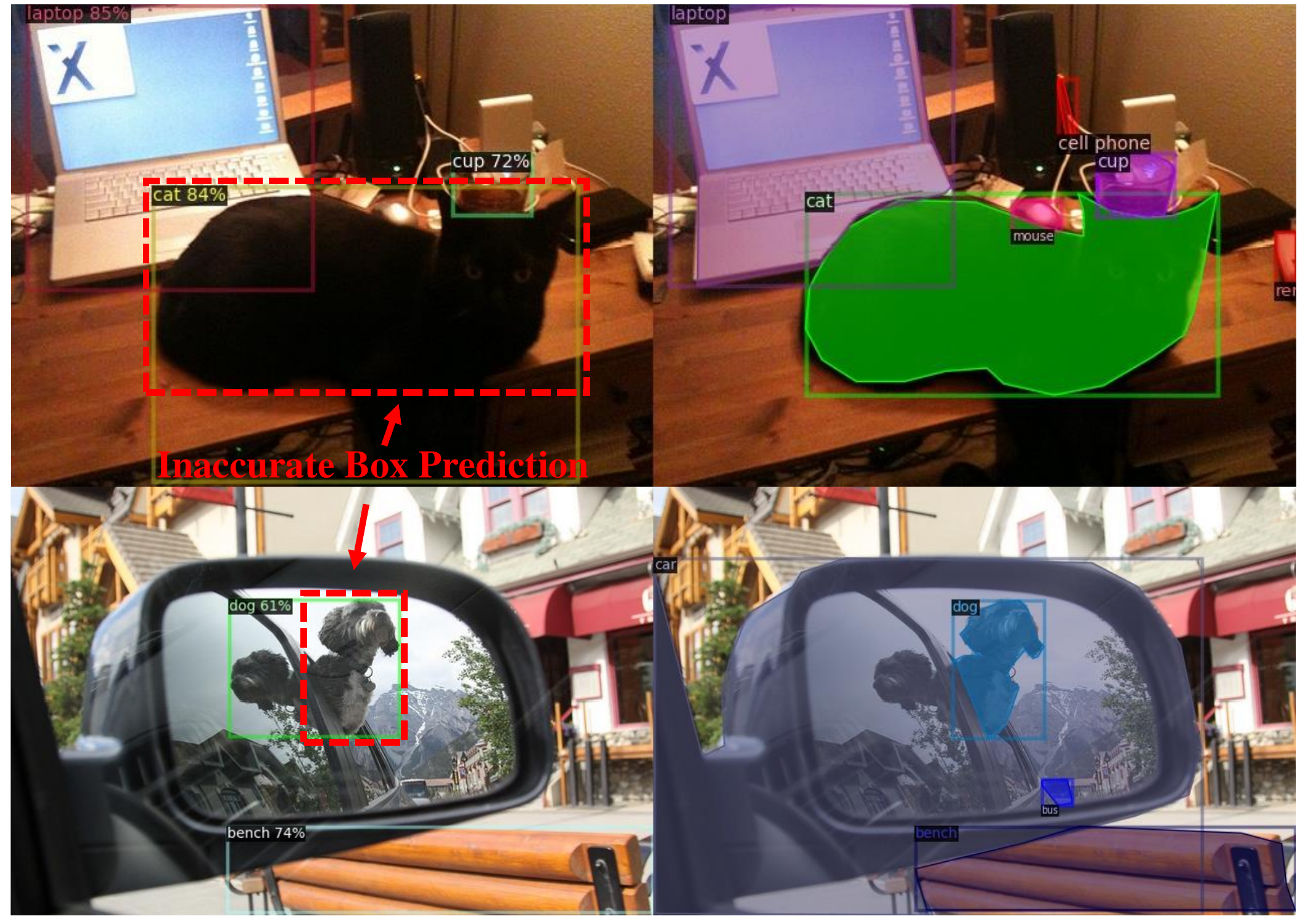}
    % \vspace{-0.1cm}
    \caption{Inaccurate box predictions for categories "cat" and "dog" (COCO Image ID: "000000119828" and "000000263463")
    }
    \label{fig:wrong_predict_location}
    % \vspace{-0.4cm}
\end{figure}

\paragraph{Visualization over special cases.} We further visualize some prediction results of in Fig.~\ref{fig:missing_prediction} and Fig.~\ref{fig:wrong_predict_location}. Although the performance of the model is already high, there are still instances of missed predictions and inaccurate predictions, as shown in these figures. It remains challenging to predict all objects completely and accurately.


\subsection{Analysis of Annotation Quality}
\begin{figure}[h]
    \centering
    % \vspace{-0.2cm}
    \includegraphics[width=0.8\linewidth, bb=0 0 800 300]{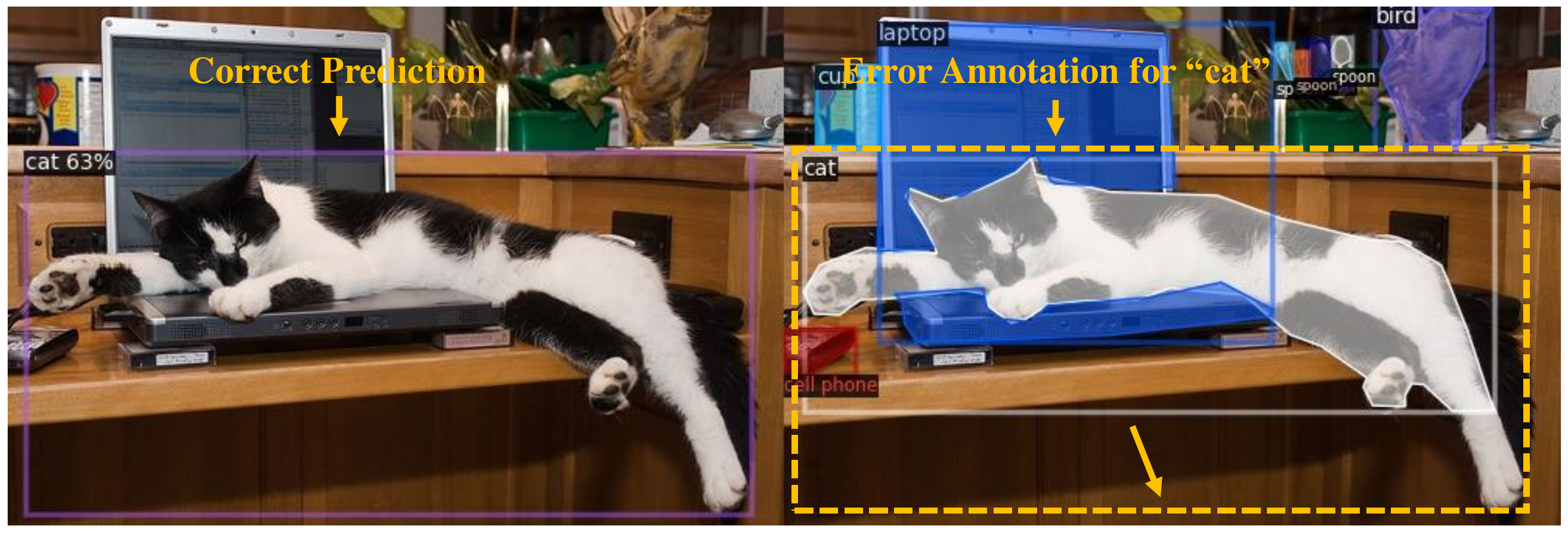}
    % \vspace{-0.1cm}
    \caption{Error annotations in the data. The left is prediction and the right is annotation. (COCO Image ID: "000000119233")
    }
    \label{fig:wrong_labels}
    % \vspace{-0.4cm}
\end{figure}

In this section, we compare the visualization of the model's predicted results with the annotated ground truths in COCO. In all the figures presented, the left half represents the model's predicted results, while the corresponding annotations from the dataset are shown on the right half. Then we can identify several issues regarding the dataset annotations:

\paragraph{Error annotations.} As shown in Fig.~\ref{fig:wrong_labels}, regarding the annotation for "cat" in the figure, the model makes a correct prediction, whereas the annotation in the dataset is incorrect in terms of its location. This could further impact the evaluation results of the model.

\begin{figure}[h]
    \centering
    % \vspace{-0.2cm}
    \includegraphics[width=0.8\linewidth, bb=0 0 650 200]{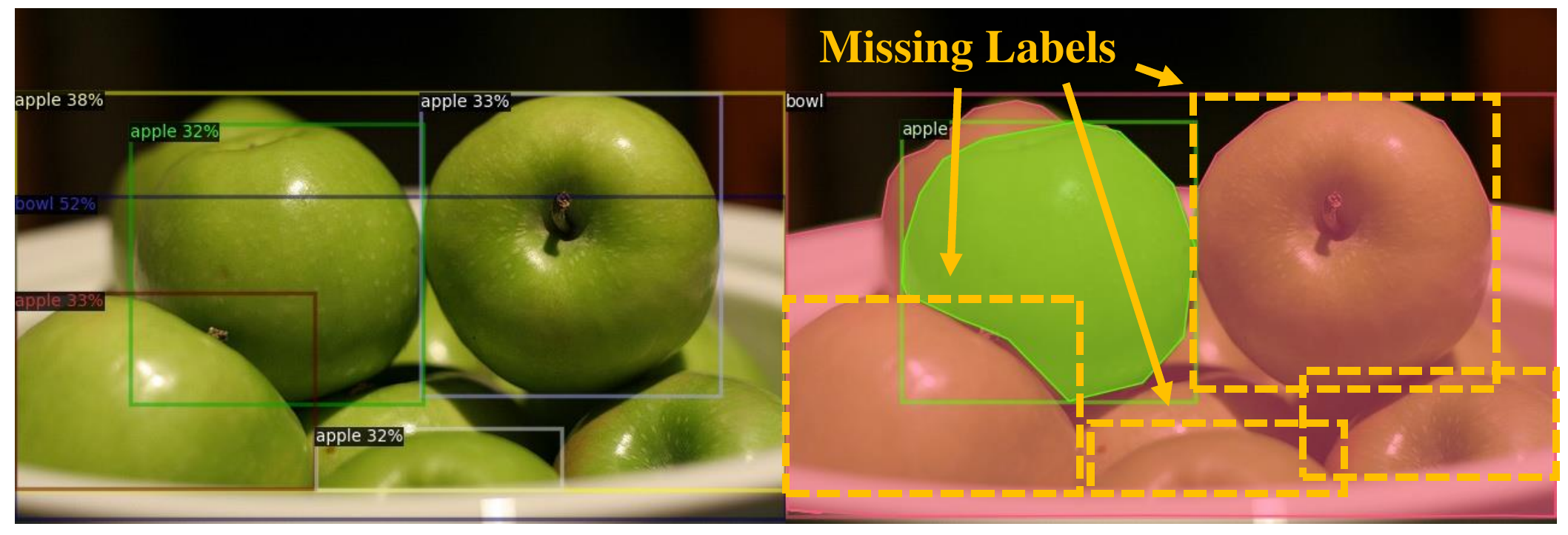}
    % \vspace{-0.1cm}
    \caption{Missing labels in the data. The left is prediction and the right is annotation. (COCO Image ID: "000000002149")
    }
    \label{fig:missing_labels}
    % \vspace{-0.4cm}
\end{figure}

\paragraph{Missing labels.} As shown in Fig.~\ref{fig:missing_labels}, the model accurately predicts the position and category of each object (``apple''). However, in the corresponding dataset annotations, only one apple is labeled, and the remaining apples and bowl are labeled as the same category "bowl", resulting in missing annotations for the apples and an incorrect annotation for the position of the bowl.

\begin{figure}[h]
    \centering
    % \vspace{-0.2cm}
    \includegraphics[width=0.8\linewidth, bb=0 0 500 250]{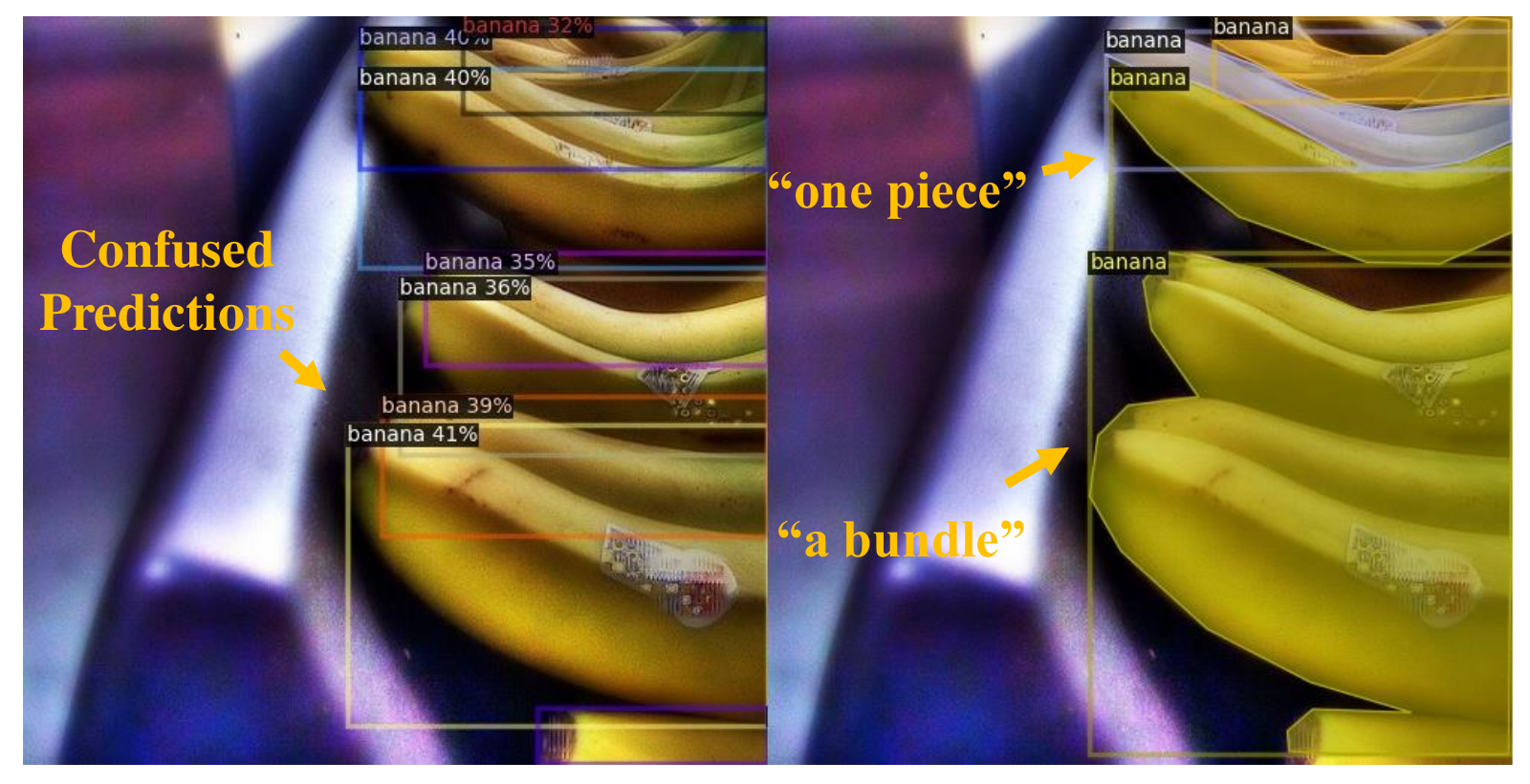}
    % \vspace{-0.1cm}
    \caption{Inconsistency annotation standards in the data. The left is prediction and the right is annotation. (COCO Image ID: "000000028809")
    }
    \label{fig:inconsistency_annotation_standards}
    % \vspace{-0.4cm}
\end{figure}

\paragraph{Inconsistency annotation standards.} As illustrated in Fig.~\ref{fig:inconsistency_annotation_standards}, there exist inconsistent labeling standards in the dataset annotations. For instance, in the same image, some annotations label the "banana" category as a bundle, while others more accurately label each individual banana as a separate object. We can observe a significant inconsistency in annotation standards in the dataset by comparing the prediction results and annotations. This inconsistency may lead to the following challenges during the model training process: (1) an inability to learn the correct data distribution, subsequently affecting the generalization capabilities of the model. (2) difficulties in determining the accurate location of the  predicted bounding boxes for such categories. (3) unstable training, which may impact the convergence of the model.

The aforementioned annotation issues may potentially impact the performance and generalization capability of the model. Hence, in our future research, it is \textbf{essential to not only focus on the performance of the models but also improve the quality of the dataset annotations}.

% \paragraph{Analysis of Annotation Quality}

% 通过可视化预测结果与ground truth，分析现有数据集的缺陷以及模型的缺陷，呼吁需要更多高质量的标注数据

% 4.1 + 4.2 结合起来讲，引出标注质量问题

% \subsection{Analysis of the Main Performance Improvements of Large Models.}

% 分析大模型与小模型之间性能的主要差异点

% 对未来的展望: 1. 更高质量的标注数据，更高效的标注流程 (human-interactive) 2. 结合最新的技术，例如GPT-4 3. 从close-set到open-set, detect anything （可以作为一个单独的章节去讲）

% sec 4.3 单独成章节

\section{Visualizations of Model Predictions}
\begin{figure}[h]
    \centering
    % \vspace{-0.2cm}
    \includegraphics[width=0.8\linewidth, bb=0 0 500 800]{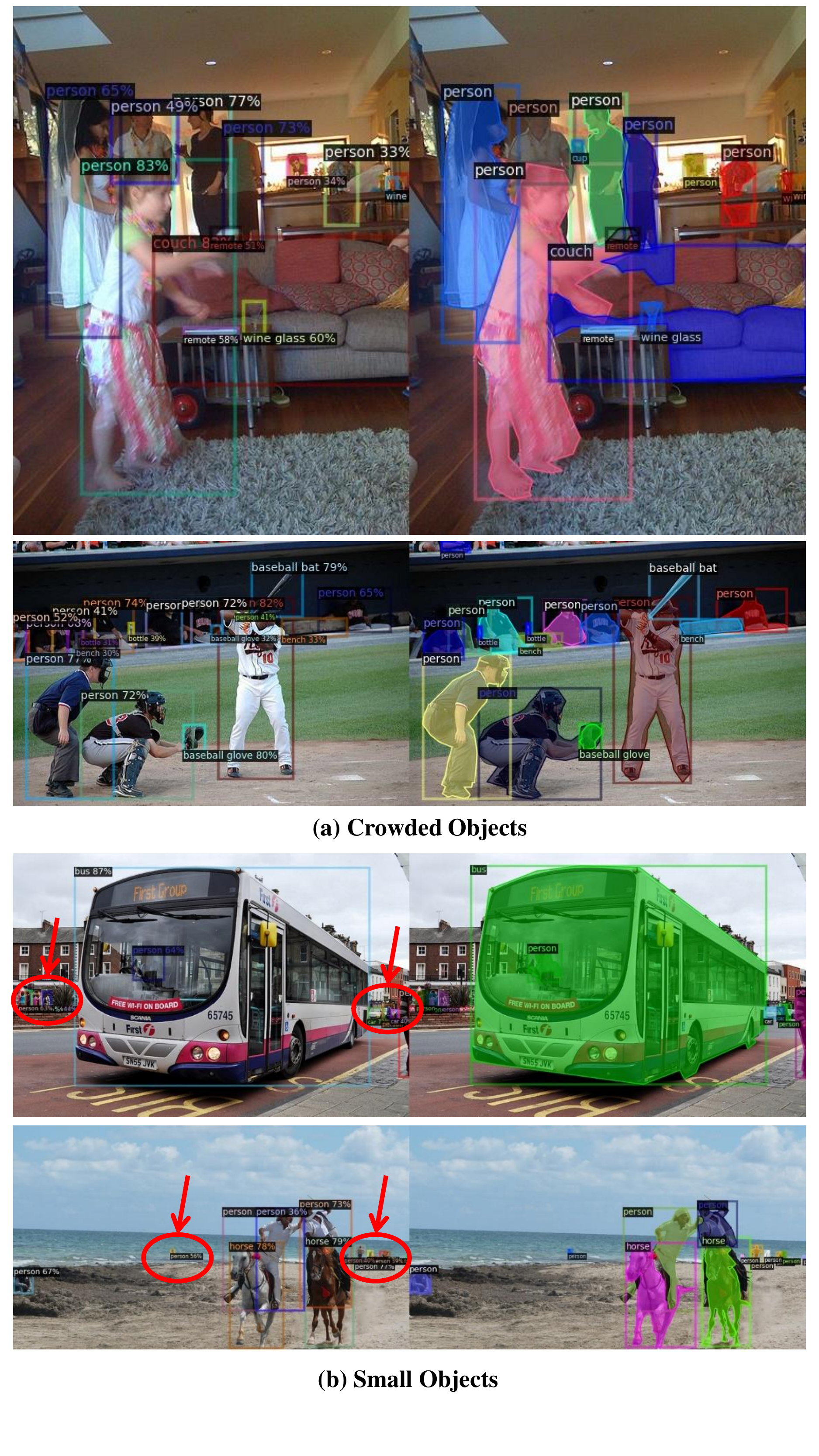}
    % \includegraphics[width=1.0\linewidth]{resources/good_cases_2.pdf}
    % \vspace{-0.1cm}
    \caption{Visualizations of model predictions. We present two hard cases: dense objects and small objects (marked in red arrows) in (a) and (b), respectively. We present the model predictions on the left and the grounded truths on the right. (COCO Image ID: "000000003934", "000000006471", "000000005037" and "000000007281")
    }
    \label{fig:vis1}
    % \vspace{-0.4cm}
\end{figure}

\begin{figure}[h]
    \centering
    % \vspace{-0.2cm}
    % \includegraphics[width=1.0\linewidth]{resources/good_cases_1.pdf}
    \includegraphics[width=0.8\linewidth, bb=0 0 500 700]{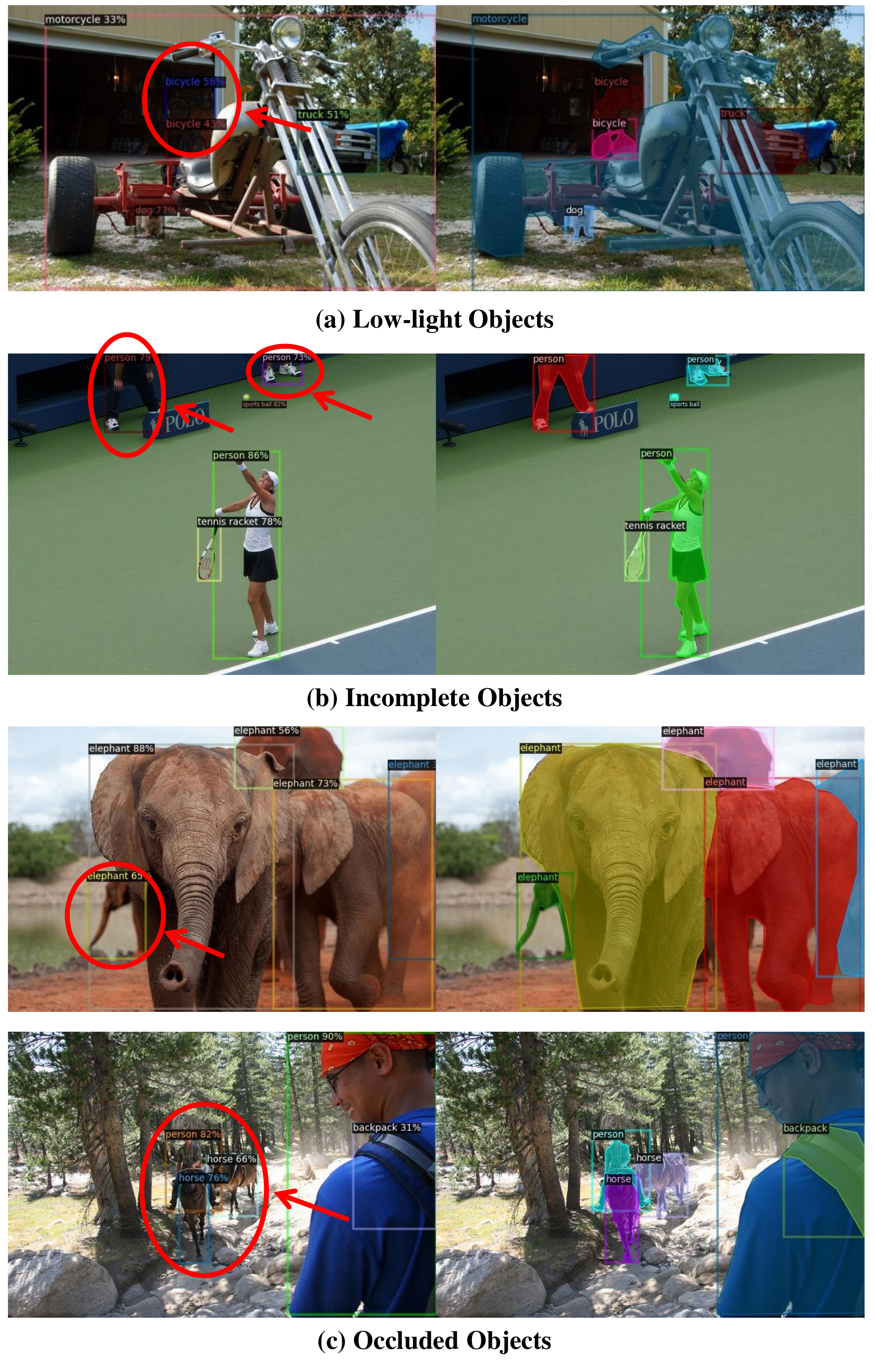}
    % \vspace{-0.1cm}
    \caption{Visualizations of model predictions (continued). We present three hard cases: low-light objects, incomplete objects,  and small objects in (a), (b), and (c), respectively. The hard objects are marked in red arrows. We present the model predictions on the left and the grounded truths on the right. (COCO Image ID: "000000007386", "000000037988", "000000007108" and "000000023034")
    }
    \label{fig:vis2}
    % \vspace{-0.4cm}
\end{figure}

Here we provide more visualization results of our Focal-Stable-DINO in Fig.~\ref{fig:vis1} and \ref{fig:vis2}. We select some challenging cases to present the superior performance of our model.

In Fig.~\ref{fig:vis1}, we show two hard cases involving crowded objects and small objects, which are highlighted with red arrows. We present the model predictions on the left and the ground truths on the right. We observe that Focal-Stable-DINO successfully predicts the locations and types of these difficult objects, demonstrating its robustness in handling complex image scenarios.
Similarly, in Fig. \ref{fig:vis2}, we show three hard cases - low-light objects, incomplete objects, and small objects, which are also marked with red arrows. Despite these challenging conditions, our model makes accurate predictions for these objects, demonstrating its ability to tackle various real-world situations.
These visualizations provide additional evidence of the effectiveness of our Focal-Stable-DINO model. Our model's ability to detect difficult objects in diverse scenarios shows its robustness and highlights its potential for using in more challenging scenarios.

\section{Conclusion}

In conclusion, this study aims to develop a strong and reproducible object detection model using only publicly available resources. We accomplish this by combining a Focal-Huge backbone with a Stable-DINO detector to create the Focal-Stable-DINO model. Focal-Stable-DINO achieves the best result among models trained with public datasets, without relying on complex training methods or data processing techniques. We hope that our study provides a compelling alternative to the current trend of relying on large-scale private data and complicated training techniques. 

Our evaluation of this model on the COCO dataset reveals several insights into object detection performance. We find a significant performance disparity across different object classes and further observe that there is still significant room for improvement in detecting small objects. Additionally, we identify inconsistencies and inaccuracies in the COCO annotations, highlighting the importance of constructing accurate and consistent annotations for reliable assessments of object detection models.

\small{\textit{"Like a rose \cite{zhang2023rose}, each person can unfold their own unique beauty. May we witness a flourishing of varied approaches and techniques in this field."}}

\clearpage
\bibliographystyle{plain}
\bibliography{neurips_2022}

%%%%%%%%%%%%%%%%%%%%%%%%%%%%%%%%%%%%%%%%%%%%%%%%%%%%%%%%%%%%

% \appendix

% \section{Visualization}

% \input{images_tex/vis}

\end{document}